\documentclass{article}

\usepackage[final, nonatbib]{nips_2018}

\usepackage[utf8]{inputenc} 
\usepackage[T1]{fontenc}    
\usepackage{hyperref}       
\usepackage{url}            
\usepackage{booktabs}       
\usepackage{amsfonts, amsmath, amssymb}       
\usepackage{nicefrac}       
\usepackage{microtype}      
\usepackage{graphicx}
\usepackage{enumitem}

\usepackage{color}

\title{Clinical Concept Extraction with Contextual Word Embedding}

\author{
  Henghui~Zhu   \\
  Boston University\\
  \texttt{henghuiz@bu.edu} \\
  \AND
  Ioannis Ch.~Paschalidis \\
  Boston University\\
  \texttt{yannisp@bu.edu} \\
   \And
   Amir Tahmasebi \\
   Philips Research North America \\
   \texttt{Amir.Tahmasebi@philips.com} \\
}

\begin{document}

\maketitle

\begin{abstract}
Automatic extraction of clinical concepts is an essential step for turning the unstructured data within a clinical note into structured and actionable information. 
 In this work, we propose a clinical concept extraction model for automatic annotation of clinical problems, treatments and tests in clinical notes utilizing domain-specific contextual word embedding. A contextual word embedding model is first trained on a corpus with a mixture of clinical reports and relevant Wikipedia pages in the clinical domain. Next, a bidirectional LSTM-CRF model is trained for clinical concept extraction using the contextual word embedding model. We tested our proposed model on the I2B2 2010 challenge dataset. Our proposed model achieved the best performance among reported baseline models and outperformed the state-of-the-art models by 3.4\% in terms of F1-score. \footnote{Codes are available at \url{https://github.com/noc-lab/clinical_concept_extraction} }
\end{abstract}


\section{Introduction}

Automatic clinical concept extraction is a crucial step in transforming the abandon of unstructured patient clinical data into a set of actionable information. A major challenge towards developing clinical Named Entity Recognition (NER) tools is access to a corpus of labeled data. The 2010i2b2/VA challenge \cite{uzuner20112010} released such corpus of annotated clinical notes to facilitate the development of clinical concept extraction systems. Since the release of the 2010 i2b2/VA dataset, there have been numerous efforts on developing NER tools reported in the literature. 
In earlier works, Conditional Random Fields (CRF) with token-level features were proposed for concept extraction  \cite{de2011machine,jonnalagadda2012enhancing,fu2014improving}. Nevertheless, such approaches require manual feature engineering, which makes its application limited. Recently, with the advancement of deep learning models and its usage for natural language processing, Recurrent Neural Network (RNN) models, such as Long Short-Term Memory (LSTM), have been widely used for deriving contextual features for CRF model training and have demonstrated promising performance for clinical concept extraction tasks \cite{boag2015cliner,chalapathy2016bidirectional,boag2018cliner,Xu-2018}. Although RNN models provide a good set of features for NER, they depend heavily on the word embedding models \cite{mikolov2013distributed,pennington2014glove}, which are not good at dealing with the complex characteristics of word use under different linguistic contexts.

Recently, Peters et al. proposed a contextual word-embedding model (ELMo), which claims to improve the performance of various NLP tasks such as sentiment analysis, question answering and sequence labeling \cite{peters2018deep}. By including the features of a language model, the word representations of the ELMo contain richer information compared to a standard word embedding such as skip-gram \cite{mikolov2013distributed} or Glove \cite{pennington2014glove}. Although the ELMo model is shown to have a good performance in some NER tasks such as the CoNLL 2003 NER task, it is trained on a corpus in a general domain \cite{chelba2013one} and as a result, does not demonstrate a desired performance for a clinical concept extraction task. This could be due to the fact that clinical text is structured differently compared to text in a general domain and therefore, the language model trained on a general domain corpus fails to generalize well on it.

In this work, we train an ELMo model in a corpus with a mixture of clinical reports and relevant Wikipedia pages in the clinical domain. Next, a bidirectional LSTM-CRF model is applied to identify the clinical concepts. This model is trained and tested on the 2010 i2b2/VA challenge \cite{uzuner20112010}. Our proposed model achieves the best performance among others the state-of-the-art models by  3.4\% in terms of F1-score.

\section{Dataset}

In this work, we used the data provided by the 2010 i2b2/VA challenge \cite{uzuner20112010} for training a clinical concept extraction system. Due to the restrictions introduced by Institutional Review Board (IRB), only a portion of data from the original dataset is available. The released dataset consists of clinical summaries from three different medical sites: Partners Healthcare, Beth Israel Deaconess Medical Center, and the University of Pittsburgh Medical Center. There are three clinical concepts annotated in this corpus: problems, tests, and treatments. There are 170 summaries for training and 256 for test. The dataset statistics are shown in Table \ref{tab:statistics}.

\begin{table}[htbp]
	\centering
	\caption{Number of reports, sentences, tokens and named entities in training and test corpus.}\label{tab:statistics}
	\begin{tabular}{lcccccc}
		\toprule
		 corpus & reports & sentences & tokens & problem & treatment & test\\
		 \midrule
		training & 170 & 16,414 & 149,541 & 7,073 & 4,844 & 4,606\\
		test & 256 & 27,763 & 267,249 & 12,592 & 9,344 & 9,225\\
		\bottomrule
	\end{tabular}
\end{table}

\section{Methods}

The proposed model consists of an ELMo model trained using a corpus in a clinical domain and a bidirectional LSTM-CRF model for the clinical context extraction.

\subsection{ELMo Training}

ELMo \cite{peters2018deep} is a recently proposed model that extends a word embedding model with features produced by a bidirectional language model. It has been shown that the utilization of ELMo for different NLP tasks results in improved performance compared to other types of word embedding models such as skip-gram and Glove\cite{peters2018deep,Kitaev-2018-SelfAttentive,he2018jointly}. Although \cite{peters2018deep} released several pretrained ELMo models, all of them are trained on a general text corpus \cite{chelba2013one}. We believe the clinical reports are structured differently compared to such general language corpora. Therefore, it is necessary to train an ELMo model specifically for the clinical domain in order to achieve the desired performance for clinical NLP tasks including concept extraction. To do so, the following corpora were considered for training the ELMo model.
\begin{itemize}[leftmargin=*]
	\item Wikipedia pages with titles that are items (medical concepts) in a standard clinical ontology, known as SNOMED CT \cite{rogers2008snomed}. The following sections were excluded: `See also', `References', `Further reading' and `External links'. Furthermore, if a term has more than one Wikipedia page, we exclude all these pages in the corpus for avoiding introducing ambiguity.
	\item The discharge summaries and radiology reports from MIMIC-III dataset \cite{johnson2016mimic}. Phrase normalization is performed on the deidentified entities in the dataset, e.g., converting all names to `John Does', removing the prefixes and suffixes for deidentified data and numbers. The reason for the normalization is to transform the deidentified reports to be more similar to actual reports.
\end{itemize}


\begin{table}[htbp]
	\caption{The corpus statistics for training the ELMo model.}
	\label{tab:elmo_traning}
	\centering
	\begin{tabular}{ccc}
		\toprule
		& number of sentences & average sentence length\\
		\midrule
		Wikipedia pages & 3,687,501 & 20.84 \\
		discharge summaries from MIMIC-III & 10,457,035 & 9.61 \\
		radiology reports from MIMIC-III & 12,786,115 & 9.62 \\
		\bottomrule
	\end{tabular}
\end{table}

This paper uses our in-house built sentence segmentation tool\footnote{\url{https://github.com/noc-lab/simple_sentence_segment}} to detect sentence boundary. NLTK \cite{bird2009natural} is used for tokenization. Statistics of the training corpus is shown in Table \ref{tab:elmo_traning}. For training the ELMo model, we use the default hyperparameter settings shown in \cite{peters2018deep}. In detail, a character-based Convolutional Neural Network (char-CNN) embedding layer \cite{kim2016character} is used with character embeddings dimension $16$, filter widths $[1,\,2,\,3,\,4,\,5,\,6,\,7]$ and number of filters $[32, 32, 64, 128, 256, 512, 2014]$. Next, a two-layer bidirectional LSTM (bi-LSTM) with 4,096 hidden units in each layer is considered. After each char-CNN embedding and LSTM layer, the output is projected to 512 dimensions and a high-way connection \cite{srivastava2015highway} is applied. In addition, we define the vocabulary in the language model as the tokens that appear not less than 5 times in the corpus. 

We randomly split the whole corpus into a training corpus (90\%) and a testing corpus (10\%). We train an ELMo model using the training corpus for $10$ epochs. The average perplexity in the testing corpus is $17.80$. On the other hand, the ELMo model\footnote{The `Original' model in section `Pre-trained ELMo Models' at \url{https://allennlp.org/elmo}} trained in a corpus of a general domain only achieves a perplexity of $628.26$. Despite the fact that comparing these two perplexities is not fair due to the different vocabularies in the two language models, such large gap between them suggests that training a specific ELMo model for the clinical domain is necessary.

\begin{figure}[htbp]
	\centering
	\includegraphics[width=\linewidth]{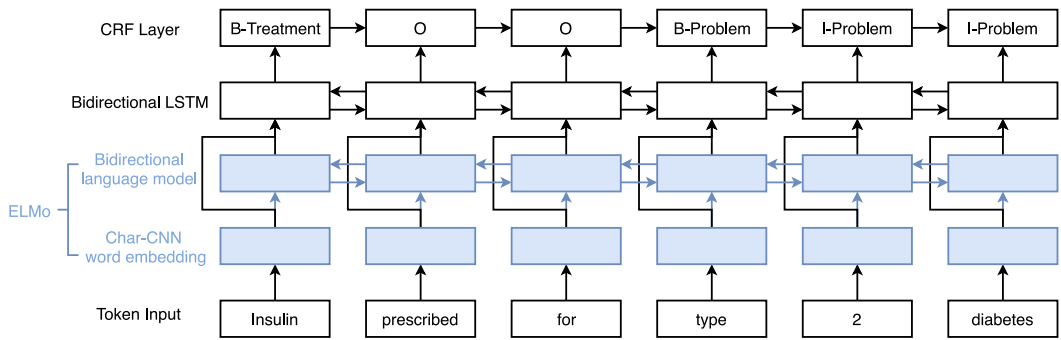}
	\caption{The proposed bidirectional LSTM model diagram for medical concept extraction. The char-based CNN word embedding and the bidirectional language model layers (ELMo), shown in the blue color, are pretrained with corpus from the clinical domain and hold fixed during training the NER model.} \label{fig:diagram}
\end{figure}

\subsection{Bidirectional LSTM CRF model for NER}

We use a bidirectional LSTM-CRF \cite{huang2015bidirectional} model for the NER task. The architecture of the proposed model is shown in Figure \ref{fig:diagram}. The input is a list of tokens. Contextual word embeddings are generated as a learnable aggregation of char-CNN word-embedding layer and two bi-LSTM layers for the language model. Suppose $\mathbf{x}_k$ is the context-independent token representation for the $k$th token in the sentence produced by the character CNN layer and denote $\mathbf{h}^0_k = [\mathbf{x}_k, \mathbf{x}_k]$ as the token layer. Also denote $\mathbf{h}^1_k = [\overleftarrow{\mathbf{h}^1_k}, \overrightarrow{\mathbf{h}^1_k}]$ and $\mathbf{h}^2_k = [\overleftarrow{\mathbf{h}^2_k}, \overrightarrow{\mathbf{h}^2_k}]$ the two bidirectional LSTM layers in ELMo with forward and backward language models. Following \cite{peters2018deep}, we use 
\begin{equation*}
	\mathbf{y}_k  = \gamma \sum_{i=0}^{2} s_i \mathbf{h}_k^i
\end{equation*}
as the features for the NER model, where $\gamma$ is a scale factor and $s_i$'s are softmax-normalized weights. During training the NER model, the parameters of the ELMo model is fixed while $\gamma$ and $s_i$'s are learnable parameters.
Next, a two layer bidirectional LSTM is applied with $\mathbf{y}_k$'s as input. Finally, a linear-chain CRF layer \cite{huang2015bidirectional} is applied for predicting the label of each token. In this paper, the BIO-tagging format is used, as shown in Figure \ref{fig:diagram}.

\section{Results}

A two-layer bidirectional LSTM is used for NER task, each of which consists of 256 hidden states. For regularization, dropout \cite{gal2016theoretically} is applied to the LSTM layer with a rate of $0.5$. We train the model with the Adam optimizer \cite{kingma2014adam} using a learning rate of $0.001$, a batch size of 32, and 200 epochs.

\begin{table}[htbp]
	\caption{Performance comparison between proposed models and the state-of-the-art models on the 2010 i2b2/VA dataset.}
	\label{tab:result}
	\begin{tabular}{lccc}
		\toprule
		Methods & Precision & Recall & F1\\
		\midrule
		Distributional semantics CRF \cite{jonnalagadda2012enhancing} * & 85.60 & 82.00 & 83.70\\
		Hidden semi-Markov model \cite{de2011machine} * & 86.88 & 83.64 & 85.23\\
		\midrule
		Truecasing CRFSuite \cite{fu2014improving} & 80.83 & 71.47 & 75.86\\
		CliNER \cite{boag2015cliner} & 79.5 & 81.2 & 80.0\\
		Binarized neural embedding CRF \cite{wu2015study} & 85.10 & 80.60 & 82.80\\
		Glove-BiLSTM-CRF \cite{chalapathy2016bidirectional} & 84.36 & 83.41 & 83.88\\
		CliNER 2.0 \cite{boag2018cliner} & 84.0 & 83.6 & 83.8\\
		Att-BiLSTM-CRF + Transfer \cite{Xu-2018} & 86.27 & 85.15 & 85.71\\
		\midrule
		ELMo(General) + BiLSTM-CRF (Single) ** & $83.26 \pm 0.25 $ & $81.84 \pm 0.22 $ & $82.54 \pm 0.14 $ \\
		ELMo(Clinical) + BiLSTM-CRF (Single) ** & $87.44 \pm 0.27$  & $86.25 \pm 0.26$ & $86.84 \pm 0.16$\\
		ELMo(Clinical) + BiLSTM-CRF (Ensemble) & \textbf{89.34} & \textbf{87.87} & \textbf{88.60}\\
		\bottomrule
	\end{tabular}

	{\flushleft \small * These models were trained using the original larger labeled dataset of the 2010 i2b2/VA challenge. \\
    ** Performances are reported as mean $\pm$ standard deviation across 10 runs with different random seeds.}
\end{table}

Three different scenarios of the proposed ELMo-based model were considered and compared in this study: Two BiLSTM-CRF models were trained using 1) an ELMo model trained on a general domain corpus \cite{peters2018deep}, referred to as ``ELMo(General) + BiLSTM-CRF (Single)"; and 2) an ELMo model trained on a clinical corpus as described before, referred to as ``ELMo(Clinical) + BiLSTM-CRF (Single)". The training was performed 10 times starting with 10 different random seeds and the mean and standard deviation of the performance metrics were reported. Besides, we also trained an ensemble model based on the most voted label by the 10 models for each token. The performance of our models and several previously published baseline models for the 2010 i2b2/VA challenge is reported in Table \ref{tab:result} in terms of precision, recall and F1-score for exact class spans using the definition given in \cite{uzuner20112010}. As expected, it can be observed from the table that an ELMo model trained using domain-specific data results in significant improvement in performance. The BiLSTM-CRF model with ELMo trained on the clinical corpus outperforms other alternatives. Furthermore, it was observed that our best model yielded similar performance among three types of named entities (problem, treatment and test). 

%


\section{Conclusions}
Contextual word embedding approach such as ELMo has exhibited a promising performance in many natural language processing tasks. In this paper, we trained a domain-specific ELMo model using a clinical domain-specific corpus and furthermore, utilized it for building a clinical concept extraction tool. We trained and tested the proposed model using the dataset provided by the 2010 i2b2/VA challenge. To the best of our knowledge, our model yields the best performance compared to the reported prior work by a significant margin of 3.4\% in terms of F1-score. 

From the results reported in this work, one can conclude that training a domain-specific language model is essential to achieve high performance for NER tasks. Nevertheless, access to a large size domain-specific corpus is a known challenge for training a language model. In this work, we demonstrated an effective yet simple approach to create such corpus for a clinical domain by filtering a general domain corpus such as wiki-pages using a domain-specific ontology such as SNOMED CT.


\subsubsection*{Acknowledgments}
Research partially supported by the ONR under MURI N00014-16-1-2832, by the NSF under grants DMS-1664644, CNS-1645681, CCF-1527292, and IIS-1237022, by the NVIDIA Corporation with the donation of a Titan Xp GPU, and by the Center for Information and Systems Engineering. 
%

\small

\bibliographystyle{unsrt}
\bibliography{reference.bib}

\end{document}